\documentclass[a4paper]{article}
\usepackage{amsmath,amstext,amsgen,amsbsy,amsopn,amsfonts,amssymb}
\usepackage{easybmat}
\usepackage{graphics}
\usepackage[pdftex]{graphicx}
\usepackage{epsfig}
\usepackage{hyperref}
\usepackage{amsthm}
\usepackage{float}
\usepackage{bm}
\usepackage{color}
\usepackage{algorithmic}
\usepackage{algorithm}
\usepackage{quoting}
\usepackage{flushend}
\usepackage{balance}
\usepackage{multicol}
\usepackage[resetlabels]{multibib}
\usepackage{wrapfig}
\usepackage{esvect}
\usepackage{multirow}
\usepackage{array}

\newtheorem{theorem}{Theorem}[section]

\begin{document}
\title{\textbf{K-nearest Neighbor Search by \\Random Projection Forests}}

\author{
Donghui Yan$^{\dag\P}$, Yingjie Wang$^{\ddag\P}$, Jin Wang$^{\ddag\P}$, \\Honggang Wang$^{\ddag\P}$, Zhenpeng Li$^{\S}$
\vspace{0.11in}\\
$^\dag$Mathematics and Data Science\vspace{0.03in}\\
$^\ddag$Department of Electrical and Computer Engineering\vspace{0.03in}\\
$^\P$University of Massachusetts Dartmouth, MA\vspace{0.09in}\\
$^\S$Department of Applied Statistics, \\[0.03in]
Dali University, Yunnan, China \\[0.03in]
}

\date{\today}
\maketitle

\begin{abstract}
\noindent
K-nearest neighbor (kNN) search has wide applications in many areas, including data mining, machine learning, 
statistics and many applied domains. Inspired by the success of ensemble methods and the flexibility of tree-based 
methodology, we propose random projection forests, {\it rpForests}, for kNN search. {\it rpForests} finds kNNs by 
aggregating results from an ensemble of random projection trees with each constructed recursively through a series 
of carefully chosen random projections. {\it rpForests} achieves a remarkable accuracy in terms of fast decay in the 
missing rate of kNNs and that of discrepancy in the kNN distances. 
{\it rpForests} has a very low computational complexity. The ensemble nature of {\it rpForests} makes it easily run in 
parallel on multicore or clustered computers; the running time is expected to be nearly inversely proportional to the 
number of cores or machines. We give theoretical insights by showing the exponential decay of the probability that
neighboring points would be separated by ensemble random projection trees when the ensemble size increases. Our 
theory can be used to refine the choice of random projections in the growth of trees, and experiments show that the 
effect is remarkable.
\end{abstract}

\section{Introduction}
\label{section:introduction}
\begin{figure}[h]
\centering
\begin{center}
\hspace{0cm}
\includegraphics[scale=0.3,clip]{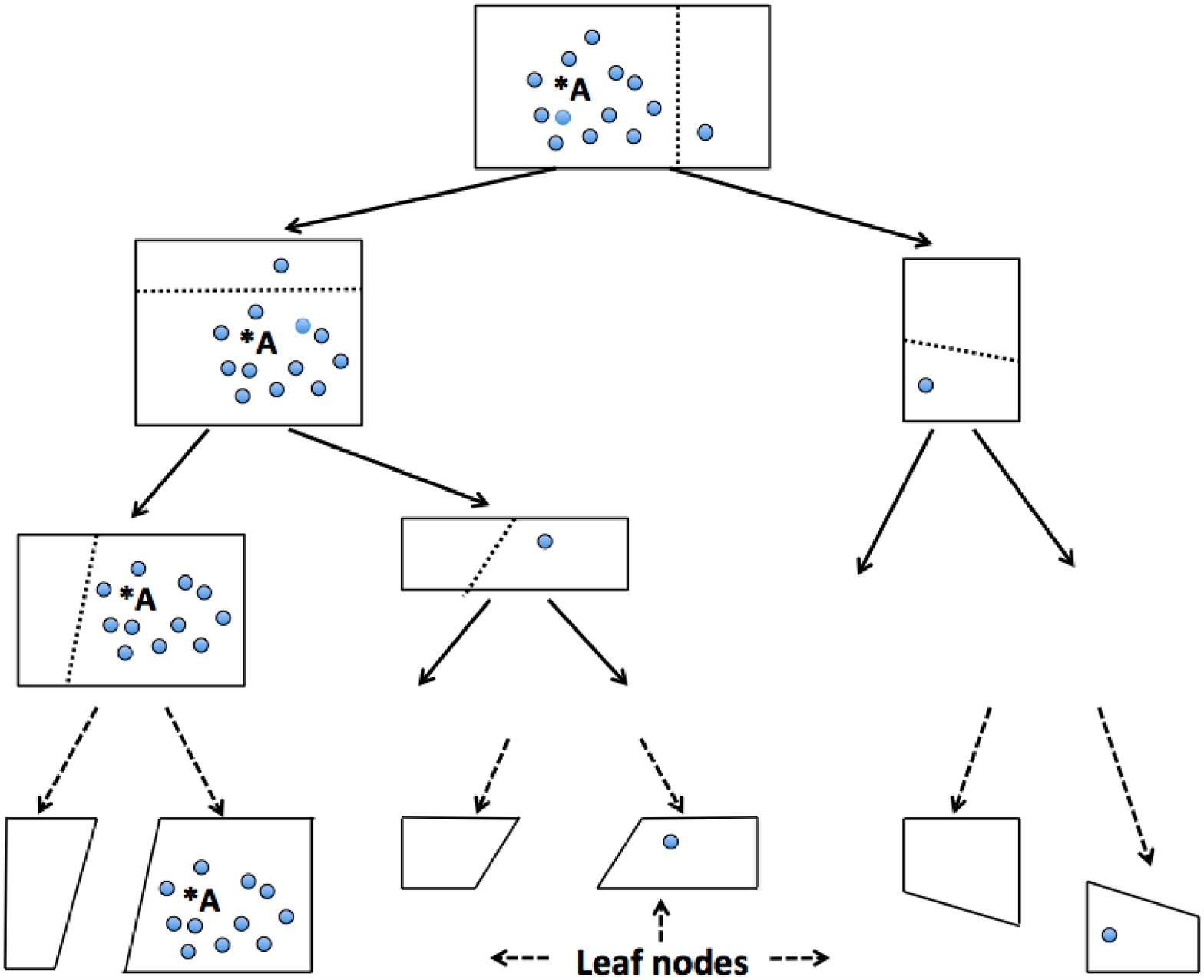}
\end{center}
\caption{\it Illustration of the near neighbors of point, $A$, and the rpTree.  } 
\label{figure:rpKNN}
\end{figure}

K-nearest neighbor (kNN) search refers to the problem of finding K points closest to a given data point on a distance metric of interest. 
It is an important task in a wide range of applications, including similarity search in data mining \cite{DongMosesLi2011,FriedmanBentleyFinkel1977}, fast kernel
methods in machine learning \cite{NystromSpectral,LucinskaWierzchon2012, ScholkopfSmola2001}, nonparametric density estimation
\cite{BickelBreiman1983,LoftsgaardenQuesenberry1965,MackRosenblatt1979} and intrinsic dimension estimation \cite{BickelYan2008,LevinaBickel2005} in statistics, as well as anomaly detection algorithms \cite{AngiulliPizzu2002,ChandolaBanerjeeKumar2007,RamaswamyRastogiShim2000}. Numerous algorithms have been proposed 
for kNN search; the readers are referred to \cite{PapadopoulosManolopoulos2005,Yianilos93datastructures} and references therein. 
Our interest is kNN search in emerging applications. Two salient features of such applications are the expected scalability of the algorithms 
and their ability to handle data of high dimensionality. Additionally, such applications often desire more accurate kNN search. For 
example, robotic route planning \cite{Kleinbort2015EfficientHM} and face-based surveillance systems \cite{Otto2017ClusteringMO}
require a high accuracy for the robust execution of tasks. However, most existing work on kNN search
\cite{LSHashing,Beygelzimer06covertrees,DasguptaSinha2015,DongMosesLi2011} have focused mainly on the fast computation 
and accuracy is of a less concern. Indeed these form the major motivations of the present work. 
\\
\\
We propose to use random projection forests ({\it rpForests}) for kNN search, inspired by the success of Random Forests (RF) \cite{RF}, 
random projection trees (rpTree) \cite{RPTree}, as well as some previous work of the author \cite{CF,YanHuangJordan2009,deepTacoma2017} 
that use the idea of ensemble or random projections. {\it rpForests} is an ensemble of rpTrees, a randomized version of the popular kd-tree \cite{Bentley1975,FriedmanBentleyFinkel1977,HuntMarkStoll2006,Otair2013}. The idea of ensemble to improve algorithmic performance is well-established in statistics and machine learning (see, for example, \cite{Bagging,RF,Adaboost,CF}), and has been the 
essential ingredient underlying some of the most successful machine learning algorithms \cite{RF,Adaboost}. More recently, the winner 
of the well-known {\it Netflix Challenge} \cite{netflix2012} is an ensemble of 106 classifiers.  
\\
\\
As {\it rpForests} uses rpTrees as its building block, it inherits several desired properties of tree-based methods. Trees 
are known to be invariant to monotonic transformations of the data and are easy to interpret. 
Tree-based methods are typically very efficient with a computational complexity 
at the order of the tree heights (which is on average the logarithm of the total number of data points). Arguably more 
importantly, tree-based methods can be viewed as recursive space partitioning \cite{Bentley1975, RPTree, YanDavis2018}. Thus data 
points living in the same tree leaf node would be ``similar". This property is frequently leveraged for approximate large scale computation, 
either with data points in the same leaf node as a large computation unit, or one of such points or a signature of them as a proxy for 
further computation \cite{DasguptaSinha2015,DhesiKar2010,LiuMooreGray2004,Sinha2014LSHVR,YanHuangJordan2009} etc. 
\\
\\
A non-desirable effect of tree-based methods is the introduction of boundary among data points not in the same leaf nodes (i.e., data 
points not in the same leaf node are viewed as lying outside the locality of each other). 
This is not a problem for some applications (e.g., \cite{YanHuangJordan2009}), as the overall decision boundary is only affected
by tree nodes near the decision boundary (which amounts to a negligible fraction of all the data points). However, this may cause 
errors to algorithms that would be
affected by the ``artificial" boundary introduced at every leaf node. For example in kNN search, it has been observed \cite{DasguptaSinha2015,LiuMooreGray2004} that the best matches for data points may be missed near the boundary of the leaf 
nodes. This is illustrated in Figure~\ref{figure:rpKNN} where rpTree is used for kNN search. Initially, the root node consists of all the 
data points. Suppose we are interested in a given data point A, and all its kNNs are marked as blue (other points are not drawn for 
clarity of visualization). The split of the root node causes one of the kNNs of point A to lie at a different child node from where A and 
all its other kNNs would lie. During the growth of rpTree, additional kNNs of A may be separated from A. Eventually, point A and most 
of its kNNs would be in the same leaf node, with a few of its kNNs landing in different leaf nodes.   
\\
\\
Several innovative ideas have been proposed as a remedy. For example, \cite{LiuMooreGray2004} 
proposed to use the spill tree where the child nodes in a k-d tree are allowed to overlap thus effectively 
reduce the chance of missing the best matches. \cite{DasguptaSinha2015} considered a variant, the 
virtual spill tree, where each data point is allocated to one leaf node but an overlapping split would route a data 
point to multiple leaf nodes and kNN search is performed on their union. The implementation 
of spill tree or its variants is, however, complicated as it involves decision on the overlapping of splits or nodes. 
\\
\\
{\it rpForests} uses the ensemble of rpTrees to reduce the chance of mis-matches in kNN search. 
For each rpTree, kNN search will be routed to one leaf node, but the search for kNNs will be 
conducted on the union of all the routed leaf nodes in the ensemble. As the growth of individual
rpTrees is independent, the union of such leaf nodes will extend the boundary beyond that of any single leaf node, thus reducing the chance of a mismatch. The ensemble nature of the algorithm and the random nature of decisions involved in the growth of rpTree make it very easy to implement and also possible to 
run the algorithm on multi-core or clustered computers. As our algorithm uses rpTree as a building block (modified in the choice of splitting direction) and resembles that of RF, we term it {\it rpForests}. 
\\
\\
Our main contributions are as follows. First, we propose a method that has the flexibility of tree-based methods and the power of ensemble methods; the method is simple to implement, highly scalable, and readily adapt to the geometry of the data. As the method is ensemble-based, easily it can run on clustered or multi-core computers. Second, we develop a theory on the probability of neighboring points being separated by ensemble rpTrees. Such a probability would explain why a tree built through random projections may be suitable for kNN search, and indeed the error decays exponentially fast when the number of trees increases. Third, our theory can be used to guide the choice of random projections---those aligning with directions along which the data stretches more are preferred.
Indeed, almost all previous methods on random trees pursue the selection of the splitting point rather than the direction (i.e., random); {\it rpForests} refines the splitting direction and our experiments suggest that this works well and such a strategy maybe applied in more general settings. 
\\
\\
The remainder of this paper is organized as follows. In Section \ref{section:method}, we give a detailed description of {\it rpForests}. This is followed by a little theory on the probability of a miss in kNN search by {\it rpForests} in Section~\ref{section:theory}. Related work are discussed in Section \ref{section:related}. In Section~\ref{section:evaluation}, we present experimental results on a wide variety of real datasets. Finally, we conclude in Section~\ref{section:conclusion}.

\section{An algorithmic description of {\it rpForests}}
\label{section:method}
{\it rpForests} uses rpTree as its building block. The growth of a tree proceeds as follows. Starting with the entire data set as the root node of a tree, it first splits the root node into two child nodes according to a splitting rule. On each of the two child nodes, it recursively applies the same procedure until some stopping criterion is met, e.g., the node becomes too small (i.e., contains too few data points). 
\\
\\
The split of a node, say $W$, in rpTree will be along a randomly generated direction, $\stackrel{\rightharpoonup}{r}$. There are many ways to randomly split a node $W$ into its left and right child,  $W= W_L \cup W_R$. 
One choice is to select a point, say $s$, uniformly at random over the interval formed by the projection for all points in $W$ onto $\stackrel{\rightharpoonup}{r}$. For a point $x\in W$, its projection onto $\stackrel{\rightharpoonup}{r}$ is given by
\begin{equation*}
\frac{r \boldsymbol{\cdot} x}{|r|^2}  \stackrel{\rightharpoonup}{r},
\end{equation*}
where $\boldsymbol{\cdot}$ indicates dot product. Define the {\it projection coefficient} of points in $W$ along direction $\stackrel{\rightharpoonup}{r}$ as $W_{\stackrel{\rightharpoonup}{r}}=\{r \boldsymbol{\cdot} x: x \in W\}$. Denote the projection coefficient of the splitting point by $c$. Then the left child $W_L$ is given by $W_L = \{x \in W: r \boldsymbol{\cdot} x <c\}$, 
and the right child $W_R$ by the rest of points. Another popular way is to choose the median of $W_{\stackrel{\rightharpoonup}{r}}$ as the split point. 
\\
\\
One advantage of rpTree over traditional tree-based methods such as the kd-tree is its ability to adapt to the geometry of the data and readily overcome the curse of dimensionality \cite{RPTree}. rpTree has been used frequently as 
a central data structure for fast computation; see, for example, \cite{DasguptaSinha2015,DhesiKar2010,YanHuangJordan2009}. 
kNN search by rpTree has a very low 
computational complexity. The growth of the tree for $n$ data points has an expected computational complexity $O(n\log(n))$, and a search involves traversing a data point from the root node down to a leaf node which, on average, costs $O(\log(n))$.  
\\
\\
The {\it rpForests} algorithm for kNN search consists of three parts---algorithm for 
rpTree (Algorithm~\ref{algorithm:neighGen}), algorithm for the selection of a splitting direction (Algorithm~\ref{algorithm:whichProj}), and algorithm to ensemble many instances of rpTrees to find kNNs (Algorithm~\ref{algorithm:kNNrpForests}). We start by describing Algorithm~\ref{algorithm:neighGen}.
\\
\\
Let $U$ denote the given data set. Let $t$ denote the rpTree to be built from $U$. Let $\mathcal{W}$ denote the set of 
working nodes. Let $n_s$ denote a constant for the minimal number of data points in a tree node for which 
we will split further. Let $P_{\stackrel{\rightharpoonup}{r}}(x)$ denote the projection coefficient of point $x$ onto line 
$\stackrel{\rightharpoonup}{r}$. Let $\mathcal{N}$ denote the set of neighborhoods s.t. each element of $\mathcal{N}$ 
is a set of neighboring points in $U$.
\begin{algorithm}
\caption{\it~~rpTree(U)}
\label{algorithm:neighGen}
\begin{algorithmic}[1]
\STATE Let $U$ be the root node of tree $t$; 
\STATE Initialize the set of working nodes $\mathcal{W} \leftarrow \{U\}$; 
\WHILE {$\mathcal{W}$ is not empty}
	\STATE Randomly pick $W \in \mathcal{W}$ and set $\mathcal{W} \leftarrow \mathcal{W} - \{W\}$; 
	\IF{$|W| < n_s$} 
		\STATE Skip to the next round of the while loop; 
	\ENDIF 
    	\STATE Generate a random direction $\stackrel{\rightharpoonup}{r}$;  
	\STATE Project points in $W$ onto $\stackrel{\rightharpoonup}{r}$, $W_{\stackrel{\rightharpoonup}{r}}=\{r \boldsymbol{\cdot} x: x \in W\}$; 
	\STATE Let $a=\min(W_{\stackrel{\rightharpoonup}{r}})$ and $b=\max(W_{\stackrel{\rightharpoonup}{r}})$; 
	\STATE Generate a splitting point $c \sim runif[a,b]$; 
	\STATE Split node $W$ by $W_L=\{x: P_{\stackrel{\rightharpoonup}{r}}(x) < c\}$ and $W_R=\{x: P_{\stackrel{\rightharpoonup}{r}}(x) \geq c\}$; 
	\STATE $W.left \leftarrow W_L$ and $W.right \leftarrow W_R$; 
	\STATE Update the working set by $\mathcal{W} \leftarrow \mathcal{W} \cup \{W_L, W_R\}$; 
\ENDWHILE
\STATE return(t); 
\end{algorithmic}
\end{algorithm} 
\\
\\
In choosing the splitting direction, a {\it basic implementation} of our algorithm would simply generate a random direction. A refinement is to generate a number, {\it nTry}, of random projections, and choose one such that the projected data stretches the most. In Statistics, the stretch of the data or the spread (dispersion) is measured by variance of the data. So we will use the standard deviation of the projected data as a measure of the data spread. As will be clear from our theory, such a choice will guide the split along a direction that the data stretches the `most' thus avoiding the situation where the data split would lead to thin slices (a setting where kNN search would easily fail). This is described as Algorithm~\ref{algorithm:whichProj}.
\begin{algorithm}
\caption{\it~~whichProjection(W, nTry)} 
\label{algorithm:whichProj}
\begin{algorithmic}[1]
\STATE Initialize $disp \leftarrow 0$, $dir \leftarrow NULL$; 
\FOR {$i=1$ to nTry}
    	\STATE Generate a random projection $\stackrel{\rightharpoonup}{r}$;  
	\STATE Project $W$ onto $\stackrel{\rightharpoonup}{r}$, and let $W_{\stackrel{\rightharpoonup}{r}}=\{r \boldsymbol{\cdot} x: x \in W\}$; 
	\STATE Let $tDisp \leftarrow SD(W_{\stackrel{\rightharpoonup}{r}})$;
	\IF{$tDisp > disp$} 
		\STATE Set $disp \leftarrow tDisp$ and $dir \leftarrow \stackrel{\rightharpoonup}{r}$;
	\ENDIF
\ENDFOR
\STATE return(dir); 
\end{algorithmic}
\end{algorithm} 
\\
 \\
Next we describe Algorithm~\ref{algorithm:kNNrpForests} for finding kNNs using the ensemble of rpTrees. Let $Q \subset U$ be the set of data points for which we wish to find their kNNs in $U$. Note that $Q$ can be the set $U$ itself.
\begin{algorithm}
\caption{\it~~kNNrpForests(Q, U)} 
\label{algorithm:kNNrpForests}
\begin{algorithmic}[1]
\FOR {$i=1$ to T}
	\STATE Build the i-th rpTree by $t_i \leftarrow rpTree(U)$; 
\ENDFOR
\FOR {each data point $q\in Q$}
	\FOR {$i=1$ to $T$}
		\STATE Let $q$ fall through tree $t_i$ and denote the leaf node that $q$ lands in by $N^{i}$; 
	\ENDFOR
	\STATE Set the consolidated neighbor set of $q$ by $N_q \leftarrow \cup_{i=1}^T N^{i}$; 
	\STATE Compute distance between $q$ and each point in $N_q$; 
	\STATE The $K$ points in $N_q$ closest to $q$ are its kNNs;
\ENDFOR
\end{algorithmic}
\end{algorithm} 
\section{Theoretical analysis}
\label{section:theory}
{\it rpForests} involves randomness in both split direction and split point in the growth of individual trees, it is desirable to know 
what performance guarantee {\it rpForests} would deliver. Our analysis will estimate the probability that two points for which one 
is a kNN of another will get separated (i.e., landing in different leaf nodes) during the growth of a tree ({\it for the basic implementation}). That is 
when there would be a miss in kNN search on a single tree. We will show that such a probability is small for any given pair of kNN points.
Of course, ensemble further reduces such a probability, and indeed the probability would decrease sharply as the ensemble size increases. 
\\
\\
\textbf{Definition.} Let $S$ be a set of points. Define its neck size, denoted as $\nu$, as the following
\begin{equation*}
\nu(S) = \inf_{\stackrel{\rightharpoonup}{r}} \sup_{x_1,x_2 \in S} \{ \mid P_ {\stackrel{\rightharpoonup}{r}}(x_1) - P_{\stackrel{\rightharpoonup}{r}}(x_2) \mid\}.
\end{equation*} 
The above definition defines the ``minor" direction of a data set, i.e., the direction along which the data points ``stretch" the least, 
while the principal direction the most. The neck is the size of the range of data points along the minor direction.
For kNN search, it is {\it undesirable} to have a small neck size during any stage of the tree growth, as that will increase 
the chance of separating two nearby points by a tree split (a potential miss in kNN search).
Algorithm~\ref{algorithm:whichProj} aims at reducing the chance of a small neck as the 
node split selected by the algorithm will be along a direction that the data ``stretches a lot".
\begin{theorem}
\label{thm:ensembleTrees}
Let $S$ be a set of data points with neck size $\nu$. Assume each tree in the ensemble $\mathcal{T}$ splits at most $J$ times, and 
the neck of the child nodes shrinks by at most a factor of $0<\gamma<1$. Then, given any two points, $A$ and $B$,  with distance $d$, 
the probability that they will be separated in the ensemble is at most
\begin{equation*}
\left( \frac{2d} {\pi \nu}  \frac{1}{\gamma^{J-2}(1-\gamma)} \right)^{|\mathcal{T}|}.
\end{equation*}
\end{theorem}
\noindent
For given $K$, the kNN distance decreases very quickly when the number of data points increases \cite{BickelYan2008,PenroseYukich2010}. 
So we can reasonably assume that $d$ is small compared to other quantities such as $\nu$ in Theorem~\ref{thm:ensembleTrees}. If one can 
properly control the value of $J$, then the probability that any given two nearby points are separated into different 
buckets (tree leaf nodes) is small for a single tree in the ensemble, and this probability will further decrease when the size of the ensemble grows. 
This is feasible as the value of $J$ only affects the size of the leaf nodes and can be adjusted.
\section{Related work}
\label{section:related}
Algorithms for fast kNN search can be divided into three categories, including hashing-, graph-, and partitioning tree-based. Hashing-based algorithms \cite{LSHashing} typically need to build a locality-sensitive hash function to find nearby points. The hash function will route neighboring points into the same hash bucket with higher probability than those points far apart. Thus, its design 
is critical and would determine the quality of kNN search. Graph-based methods \cite{NystromSpectral,LucinskaWierzchon2012} construct a kNN graph over the data points, which is then used for fast kNN search. Though such methods are generally computationally efficient, the index construction is typically slow. Space-partitioning trees are more popular for kNN search. For example, K-d tree \cite{Bentley1975} divides the data space recursively into cells along coordinate-aligning axes, and then search for kNNs by a backtrack or priority search over the tree.
Many methods have been proposed for space partitioning, such as k-means trees \cite{Muja2014ScalableNN}, cover trees \cite{Beygelzimer06covertrees}, VP trees \cite{Yianilos93datastructures} and ball trees \cite{Leibe06efficientclustering}. However, these methods typically require a long index building process for large data sets. 
\\
\\
Recently, randomized trees have been used for kNN search, such as randomized k-d tree \cite{Hartley08rOptimisedkdtree} and random projection trees \cite{DasguptaSinha2015,Hyvnen2016FastNN,Sinha2014LSHVR}. Randomized k-d trees \cite{Hartley08rOptimisedkdtree} grow a tree by randomly choosing a split point from coordinate-align axes. Random projection trees \cite{DasguptaSinha2015} choose the splitting hyperplanes sampled randomly from the unit sphere. Hyvonen et al. \cite{Hyvnen2016FastNN} uses sparse random projections to grow the tree and then ensemble. Sparse projections are generated and shared among nodes in the same level of a tree. Implementations with sparse projections are only slightly faster than dense ones, and it is not clear if there is an adverse effect to the quality of kNN search.
\\
\\
{\it rpForests} is a partitioning-tree based algorithm. It was primarily inspired by our past experience with RF and the use of rpTrees in 
algorithmic design \cite{CF,YanHuangJordan2009,deepTacoma2017}. 
Important ingredients of {\it rpForests}, the selection of random projections and the ensemble of rpTrees, are motivated by the theory we have developed. Our random projections are towards a real line (exactly as in \cite{RPTree}) which is fast and easy to implement; while those in related work are from $R^D$ to $R^d$, which involves expensive matrix multiplication and is typically slow to implement. {\it rpForests} can be viewed as complementing existing work on {\it unsupervised extensions} to RF---Cluster Forests \cite{CF} which aims at clustering by ensemble of randomized feature pursuits---in the sense that it preserves locality of nearby data points by ensemble of rpTrees. One existing work \cite{LeeYangOh2015} also uses the name {\it random projection forests}, but is fundamentally different in that it implements RF by replacing the random selection of candidate features at node splits with sparse random projections.
More recently, \cite{CanningsSamworth2017} considers random projection ensemble for classification where each ensemble instance is built on a selected random projection out of a block of projections for quality assurance. Also related is multi-view learning which combines views from different sources to improve generalization \cite{DingShaoFu2018,XuTaoXu2013}. 
\section{Experiments}
\label{section:evaluation}
We conduct experiments on a wide variety of real datasets. Most are taken from the UC Irvine Machine Learning Repository \cite{UCI}, with the exception of 
the Olivetti face from the Cambridge University Computer Laboratory (\url{http://www.cl.cam.ac.uk/research/dtg/attarchive/facedatabase.html}).
A summary of these datasets is given in Table~\ref{table:datasets}. A particularly remarkable feature about these datasets is their wide coverage of data dimensions, ranging from 19 to about 10000. We conduct experiments on both accuracy and running time, presented in Section~\ref{section:expAccuracy} and Section~\ref{section:runningTime}, respectively.
\begin{table}[h]
\begin{center}
\setlength{\extrarowheight}{2pt}
\begin{tabular}{r|rr}
\hline
Dataset                 			& Features     	&  \#Instances\\
\hline
Image Segmentation  		&19   		&2100\\
Parkinson's Telemonitoring        & 20                  & 5815\\
Wisconsin breast cancer (WDBC)            &30        &569   \\
Sensorless Drive   			& 49                  & 58509\\
Musk					& 166		&6598\\
CT Slice Localization             	& 386                & 53500\\
Smartphone Activity			&561			&7767\\
Arcene                     			& 10000		&700\\
Olivetti Face          			& 10304            & 400\\
\hline
\end{tabular}
\end{center}
\caption{A summary of datasets.} 
\label{table:datasets}
\end{table}
\subsection{Evaluation metrics}
\label{section:metrics}
Our performance evaluations are based on two metrics, the average missing rate $\overline{m}_k$ and the average discrepancy $\overline{d}_k$ 
in the k-th nearest 
neighbor distance.
Let the data be denoted by $X_1, X_2,..., X_n$. For each point $X_i$, denote the number of its
kNNs missed by the algorithm by $m(i)$. Then, 
\begin{equation*}
\overline{m}_k = \frac{1}{nk}\sum_{i=1}^n m(i).
\end{equation*}  
Clearly, for $\overline{m}_k$, the smaller the better. 
\\
\\
For each point $X_i$, 
the distance between $X_i$ and its k-th nearest neighbor is called the kNN distance for $X_i$, denoted by $d_k(i)$. Thus,
\begin{equation*}
\overline{d}_k = \frac{1}{n} \sum_{i=1}^n d_k(i).
\end{equation*}  
For any point, say $X_i$, it is clear that if one misses any of its kNNs, then the estimated kNN distance, 
$\hat{d}_k(i)$, obtained by an algorithm satisfies $\hat{d}_k(i) \geq d_k(i)$. Thus, 
the average kNN distance will be larger as well. So for $\overline{d}_k$, the smaller the better. 

\subsection{Experiments on accuracy}
\label{section:expAccuracy}
\begin{figure*}[htp]
\centering
\begin{center}
\hspace{-0.2in}
\includegraphics[scale=0.5,clip]{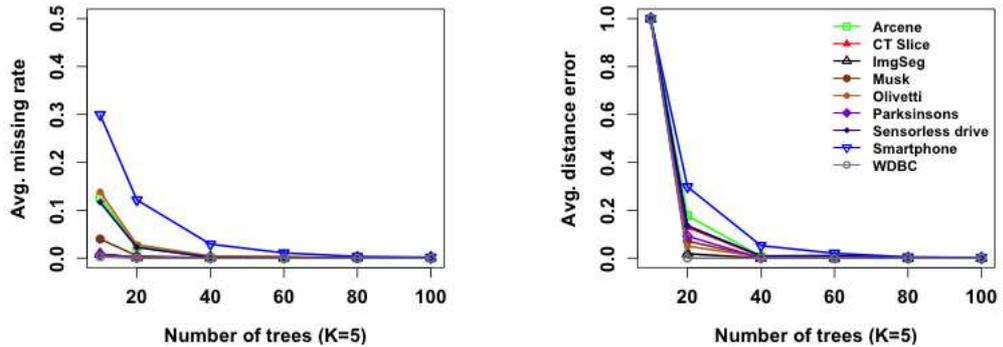}
\end{center}
\caption{\it The average missing rates and kNN distance discrepancy in kNN search when varying the number of random projection trees (K=5). } 
\label{figure:missRateDistance5}
\end{figure*}

\begin{figure*}[htp]
\hspace*{-0.18in}
\centering
\begin{center} 
\includegraphics[scale=0.5,clip]{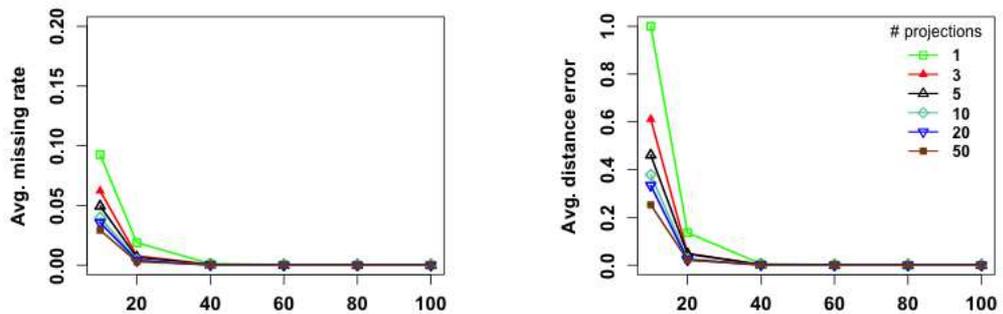}
\end{center}
\caption{\it Average missing rate and discrepancy in the kNN distance when varying the number of trees and number of random projections for the Musk dataset. } 
\label{figure:missRatevsProjs}
\end{figure*}

For each dataset, we vary the number of trees in {\it rpForests} through $10, 20, 40, 60, 80, 100$. The results are 
averaged over 100 runs. We take $K=5$ and $K=10$ (results not reported due to limit in space), 
as these two are of the most interests in real applications.
When growing the rpTrees, we fix the size of the leaf nodes (i.e., the node capacity) 
to be no more than 20 for $K=5$ and 30 for $K=10$. Here, we normalize the average discrepancy in kNN distance for each dataset 
when plotting the results. 
\\
\\
Figure~\ref{figure:missRateDistance5} shows kNN search results under the two metrics.
The number of random projections, $nTry$, are taken as $3,3,1,10,10,5,20,50,1$ for the 9 datasets 
in the order listed in Table~\ref{table:datasets}. 
It can be seen that errors, as measured by the two metrics, in kNN search by {\it rpForests} decrease quickly as 
the number of trees increases. Indeed the decay appears to be exponentially fast, as predicted by our theory. 
For most datasets we explore, and the errors sharply vanish to 0 with about 20-40 trees.
\\
\\
We also assess the role of the number of random projections, $nTry$, on kNN search. Here, we report only result ($K=5$) on the Musk 
dataset. As shown in Figure~\ref{figure:missRatevsProjs}, more trees or projections tend to improve kNN search quality.
That choosing the splitting direction from a pool of more random projections can actually be explained by our theory---more random projections will lead to a cut along a direction that the data stretches more thus reducing the chance of thin slices (a situation that would cause misses in kNN search). Similar patterns are also observed on all other datasets (not reported here due to limit in space).
\subsection{Experiments on running time}
\label{section:runningTime}
\begin{figure}[htp]
\centering
\begin{center}
\includegraphics[scale=0.6,clip]{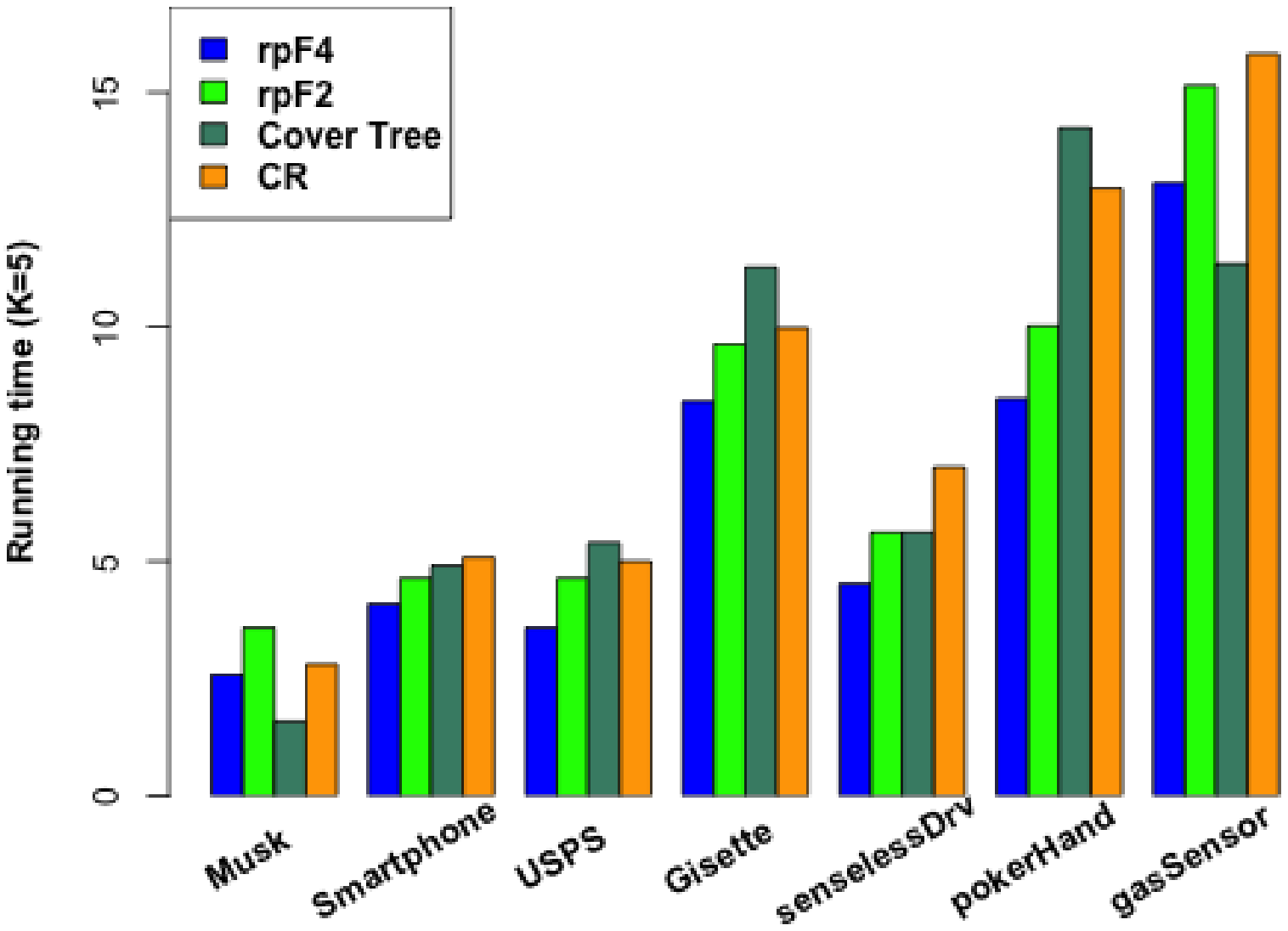}
\end{center}
\caption{\it Running time in seconds (in scale of $\log_2$) for $K=5$.} 
\label{figure:knnTime5}
\end{figure}

The running time of {\it rpForests} is evaluated and compared to the cover tree \cite{Beygelzimer06covertrees} and the CR algorithm \cite{VR2002}. We reuse some data from Table~\ref{table:datasets} with the addition of some larger datasets. This gives seven datasets with varying sizes and data dimensions; see Table~\ref{table:datasets2}. To leverage the ensemble nature of {\it rpForests} and the wide availability of multicore machines, we run {\it rpForests} with 2-core and 4-core machines, termed as rpF$_2$ and rpF$_4$, respectively. For all data, {\it rpForests} consists of 40 trees which is deemed adequate according to experiments discussed above. As the running time vary widely across different datasets, we present those in logarithmic (base 2) scale. Figure~\ref{figure:knnTime5} shows the running time on all datasets as a bar-chart for $K=5$. It can be seen that {\it rpForests} is competitive on most of the data when a 4-core machine is used. A similar pattern can be seen for $K=10$ (omitted here). The running time are produced on a MacBook Air with 1.7GHz Intel Core i7 processor and 8G memory.
\begin{table}[h]
\begin{center}
\setlength{\extrarowheight}{2pt}
\begin{tabular}{r|rr}
\hline
Dataset                 			& Features     	&  \#Instances\\
\hline
Musk                                &166                &6,598\\
Smartphone                     &561                &7,767\\
USPS digits                     &256                &11,000\\
Gisette                             &5000               &12,500\\
Sensorless Drive        		& 49                 	& 58,509\\
Poker hand					&11				&1,000,000\\
Gas sensor array		&19				&4,178,504\\
\hline
\end{tabular}
\end{center}
\caption{Datasets used for the evaluation of running time.} 
\label{table:datasets2}
\end{table}
\\
\\
For datasets we have explored, the running time decreases nearly inversely proportional to the number of cores of the 
computing machine. We hypothesize that this is true in general; a conclusive statement requires more experiments 
on machines with more cores which we leave to future work.

\section{Conclusions}
\label{section:conclusion}
{\it rpForests} is an efficient kNN search algorithm that is simple to implement, highly scalable, and readily adapt to the geometry of the underlying data. The ensemble nature of {\it rpForests} makes it easy to run in parallel on multicore or clustered computers, with running time decreasing with more cores or computers used in the computation. {\it rpForests} has the flexibility of tree-based methods; it is easy to interpret and is invariant under (monotonic) data transformations. 
On a wide variety of real datasets, with data dimension ranging from a few dozen to about 10000, {\it rpForests} quickly achieves about zero error in kNN search when the number of trees increases. We develop a theory on the fast decay of probability that neighboring points are separated by ensemble rpTrees. 
Interestingly, our theory actually gives guidance on the selection of random projections---those aligning with directions along which the data stretches more are preferred, which is confirmed by our experiments. This is different from all previous random tree-based methods that focus on the selection of the splitting point rather than the splitting direction; this strategy is expected to be applicable in more general settings. 





\bibliographystyle{plain}
%



\end{document}